\documentclass[conference]{IEEEtran}
\IEEEoverridecommandlockouts
\usepackage[nocompress]{cite}

\usepackage[dvipsnames]{xcolor}
\usepackage[colorlinks=true]{hyperref} 
\usepackage{amsmath,amssymb,amsfonts}
\usepackage{algorithmic}
\usepackage{graphicx}
\usepackage{textcomp}
\usepackage{hyperref} 
\usepackage{booktabs}
\usepackage{multirow}
\usepackage{colortbl}
\usepackage{array}
\usepackage{makecell}
\usepackage{adjustbox}
\usepackage{caption}
\usepackage{pifont}

\newcommand{\cmark}{\textcolor[rgb]{0,0.5,0}{\ding{51}}}    
\newcommand{\xmark}{\textcolor[rgb]{0.75,0,0}{\ding{55}}}   

\def\BibTeX{{\rm B\kern-.05em{\sc i\kern-.025em b}\kern-.08em
    T\kern-.1667em\lower.7ex\hbox{E}\kern-.125emX}}
\begin{document}

\title{Efficient Spatial-Temporal Focal Adapter with SSM for Temporal Action Detection
\thanks{This work was supported by JST CRONOS~JPMJCS24K4.}
}

\author{
\IEEEauthorblockN{1\textsuperscript{st} YICHENG QIU}
\IEEEauthorblockA{\textit{Department of Informatics} \\
\textit{The University of Electro-Communications}\\
Chofu, Tokyo, Japan \\
qiu-y@mm.inf.uec.ac.jp}
\and
\IEEEauthorblockN{2\textsuperscript{nd} Keiji Yanai*}
\IEEEauthorblockA{\textit{Department of Informatics} \\
\textit{The University of Electro-Communications}\\
Chofu, Tokyo, Japan \\
yanai@cs.uec.ac.jp}
}

\maketitle
\renewcommand{\thefootnote}{\fnsymbol{footnote}}
\footnotetext[0]{* Corresponding author.}
\footnotetext[0]{This paper is accepted at ICME 2026. \textcopyright~2026 IEEE. Personal use of this material is permitted. Permission from IEEE must be obtained for all other uses, in any current or future media, including reprinting/republishing this material for advertising or promotional purposes, creating new collective works, for resale or redistribution to servers or lists, or reuse of any copyrighted component of this work in other works.}

\begin{abstract}
Temporal human action detection aims to identify and localize action segments within untrimmed videos, serving as a pivotal task in video understanding. Despite the progress achieved by prior architectures like CNN and Transformer models, these continue to struggle with feature redundancy and degraded global dependency modeling capabilities when applied to long video sequences. These limitations severely constrain their scalability in real-world video analysis. State Space Models (SSMs) offer a promising alternative with linear long-term modeling and robust global temporal reasoning capabilities. Rethinking the application of SSMs in temporal modeling, this research constructs a novel framework for video human action detection. Specifically, we introduce the Efficient Spatial-Temporal Focal (ESTF) Adapter into the pre-trained layers. This module integrates the advantages of our proposed Temporal Boundary-aware SSM(TB-SSM) for temporal feature modeling with efficient processing of spatial features. We perform comprehensive and quantitative analyses across multiple benchmarks, comparing our proposed method against previous SSM-based and other structural methods. Extensive experiments demonstrate that our improved strategy significantly enhances both localization performance and robustness, validating the effectiveness of our proposed method.
\end{abstract}


\begin{figure*}[t]
  \centering
  \captionsetup{font=scriptsize}
  \includegraphics[width=0.93\linewidth]{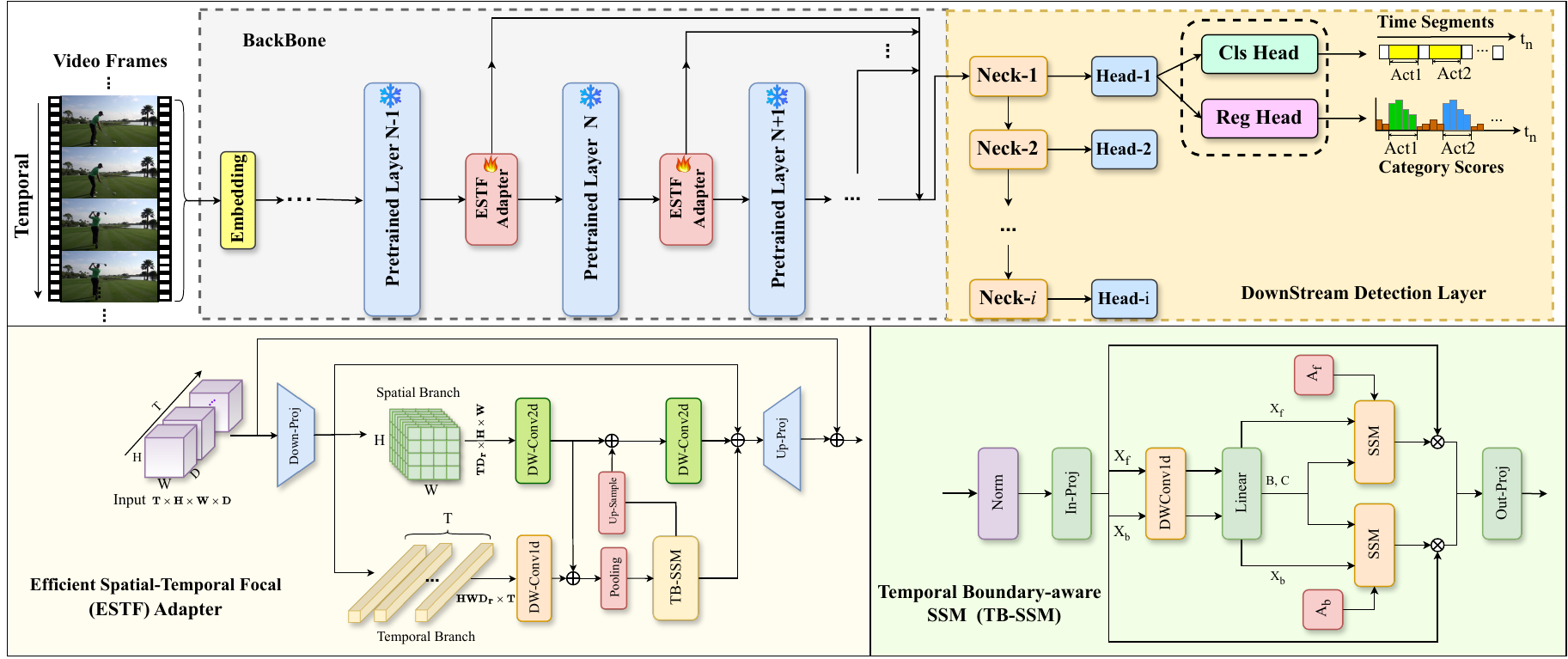}
  \caption{
    The architecture of the proposed TAD framework. We integrate ESTF Adapters into the frozen pre-trained backbone layers to adapt representations for temporal action detection. For each adapter, features are processed by two branches: (1) The Spatial Branch that extracts localized structural cues using convolutions, and (2) The Temporal Branch that employs the TB-SSM to capture asymmetric state dynamics for precise boundary regression. Finally, fused features are fed into the detection head.
    }
  \label{fig:overview}
\end{figure*}

\section{Introduction}
\label{sec:intro}
Temporal Action Detection (TAD) constitutes a fundamental and challenging research area within the broader field of video understanding. The primary objective of TAD is to precisely categorize and localize action instances within untrimmed video streams by determining their start and end timestamps. This capability is pivotal for a wide range of real-world applications, including intelligent video surveillance\cite{yang2024video,hamann2024low, geng2024uniav,qiu2025fire}, complex human behavior analysis\cite{xarles2023astra,qiu2024multipath, qiu2024efficient}, and embodied intelligence\cite{cheng2024videgothink, zhao2025urbanvideo, suglia2024alanavlm}. \par

In previous research, diverse model architectures have been developed to tackle the TAD task. Earlier works relied on multi-scale feature pyramids\cite{shi2023tridet} and Graph Convolutional Networks (GCNs)\cite{xu2020g} to capture temporal contexts. More recently, Transformer-based architectures\cite{zhang2022actionformer} have set new benchmarks by leveraging self-attention mechanisms to model global dependencies. However, these existing paradigms face inherent limitations when scaling to long video sequences. CNN-based methods often struggle with limited receptive fields, leading to insufficient modeling of long-term dependencies. Conversely, while Transformers enable global context awareness, their quadratic computational complexity $O(N^2)$ creates severe memory bottlenecks and redundancy, making them inefficient for processing high-resolution, untrimmed videos. Consequently, balancing global temporal reasoning with computational efficiency remains an unresolved bottleneck in current TAD research. Recently, the State Space Model (SSM) architecture\cite{gu2023mamba, dao2024transformers}, has emerged as a competitive paradigm, offering linear computational complexity $O(N)$ coupled with robust long-sequence modeling capabilities. This presents a significant opportunity to overcome the scalability constraints of Transformers. However, directly applying SSMs to video understanding is not trivial. Purely global modeling in SSMs can sometimes lead to the oversmoothing\cite{behrouz2024graph} issue, causing fine-grained boundary details essential for precise action localization to be submerged by global features. Furthermore, retraining large scale video backbones from scratch is computationally expensive. Therefore, a parameter-efficient architecture\cite{houlsby2019parameter} is needed to leverage Mamba’s linear efficiency while retaining discriminative boundary cues for accurate boundary detection.\par

In this paper, we propose a new TAD framework centered on the Efficient Spatial-Temporal Focal (ESTF) Adapter. Compared with full fine-tuning on pretrained model, our method integrates a spatial feature processing module and a temporal SSM module into a pre-trained video backbone. ESTF improves feature processing efficiency by decoupling spatial and temporal modeling and then fusing them through feature interaction. Specifically, it first applies spatial downsampling to reduce redundant information, and then uses a parameter-efficient Temporal Boundary-aware SSM (TB-SSM) as the core temporal component. We adopt the selective scan mechanism of SSMs to capture long-range dependencies with linear complexity, and further introduce spatial and temporal focal designs to highlight key boundary changes of video human action features. It helps reduce the oversmoothing issue and enables more robust action representation learning under complex backgrounds.
We conduct extensive comparative experiments on multiple benchmarks. The results show that, compared with prior SSM-based and Transformer-based methods, our approach consistently improves localization performance and robustness.\par

The main contributions of this paper are summarized as follows:
\begin{itemize}
\item We propose a TAD detection framework equipped with the Efficient Spatial-Temporal Focal (ESTF) Adapter, which integrates the linear-time Mamba architecture into the TAD pipeline and effectively addresses the trade-off between global modeling capability and computational cost.
\item Within the ESTF Adapter, we decouple spatial and temporal modeling and perform feature interaction and fusion, enabling the SSM module to better capture long-range dependencies while preserving fine-grained boundary information.
\item We propose Temporal Boundary-aware SSM(TB-SSM) structure, where the state transition matrices for the forward and backward directions are parameterized separately to better model the inherent asymmetry of action boundaries.
\end{itemize}

\section{Related Work}

\subsection{Temporal Action Detection Methods}
ActionFormer\cite{zhang2022actionformer} is built on the transformer architecture. TriDet\cite{shi2023tridet}  is constructed based on the multi-scale feature pyramid. TemporalMaxer\cite{tang2023temporalmaxer} uses max-pooling to extract information from the video. ActionMamba\cite{chen2024video} utilizes the SSM\cite{gu2023mamba} architecture for global extraction of temporal features. BMN\cite{lin2019bmn} introduces a Boundary-Matching mechanism to estimate temporal boundaries. G-TAD\cite{xu2020g} employs graph convolutional networks to video actions.  AFSD\cite{lin2021learning} proposes an anchor-free method. TadTR\cite{liu2022end} presents an end-to-end model based on the transformer structure, and ETAD\cite{liu2023etad} introduces sequentialized video encoding and gradient updating to reduce computational resources. 

\subsection{State Space Models of Video Understanding}
State Space Model(SSM)\cite{gu2023mamba} is a foundational approach to general sequence modeling. Based on SSM, the subsequent work VisionMamba\cite{zhu2024vision}, extended the Mamba architecture to the field of Computer Vision. VideoMamba\cite{li2025videomamba} further applies this architecture to video understanding. ActionMamba\cite{chen2024video} adopts the bidirectional parameter-sharing strategy, which reduces computational overhead. MS-Temba\cite{sinha2025ms} employs the multi-scale temporal State Space Model (SSM) architecture for video action detection task.

\section{Methodology}
This section provides an overview of our proposed TAD framework and presents the internal design details of our core Efficient Spatial-Temporal Focal (ESTF) module. We introduce the overall framework of the video action detection task constructed in this paper, as shown in Figure\ref{fig:overview}.
We first define the task object setting. Given an untrimmed video $\mathbf{V} \in  \mathbb{R}^{T \times H \times W \times D}$, where $T$, $H$, $W$, and $D$ denote the number of frames, height, width, and channel dimension respectively. The TAD task aims to accurately and efficiently predict a set of action instances $\Psi_p = \{ \phi_i = (t_{s}^{i}, t_e^i, c^i, s^i)\}_{i=1}^{M}$, where $M$ is the number of predicted action instances, and $t_s^i$, $t_e^i$, $c^i$, and $s^i$ denote the start time, end time, category, and confidence score of the i-th action instance, respectively. The ground-truth action instances are represented as $\Psi_g = \{\hat{\phi_i} = (\hat{t}_{s}^{i}, \hat{t}_e^i, \hat{c}^i)\}_{i=1}^{N}$, where $N$ is the number of ground-truth action instances.
The input video is divided into spatio-temporal patches and embedded as $\mathbf{X} = \text{PatchEmbed}(\mathbf{V}) + \mathbf{E}_{pos}$
where $\mathbf{E}_{pos} \in \mathbb{R}^{T \times D}$ denotes the positional embedding and $\mathbf{X} \in \mathbb{R}^{N \times D}$. $N$ is the number of patches as $N = T/t_p \times H/h_p \times W/w_p$, $t_p, h_p, w_p$ represent the patch size. 

We adopt the pretrained backbone with adapter modules.
Given an embedded patch sequence $\mathbf{X} \in \mathbb{R}^{N \times D}$, the video backbone consists of a stack of $L$ pretrained main blocks. The main block is augmented with an adapter module. For the $l$-th block, the feature propagation is defined as
\begin{align}
\mathbf{X}'_l &= \mathbf{X}_{l-1} + \mathbf{W}_l\big(\mathrm{LN}(\mathbf{X}_{l-1})\big), \\
\mathbf{X}_l &= \mathbf{X}'_l + \mathcal{A}_l\big(\mathbf{X}'_l\big),
\end{align}
where $\mathbf{W}_l(\cdot)$ denotes the pretrained main network.
$\mathcal{A}_l(\cdot)$ represents the adapter module, which is responsible for efficient temporal feature refinement.\par

The Neck layer is introduced to transform the backbone features into a temporally organized representation suitable for detection. Temporal downsampling is applied to aggregate temporal information and generate compact temporal feature sequences. At the $i$-th Neck layer, the temporal features are obtained as:
\begin{equation}
\mathbf{F_i} = \mathrm{TemporalPool}(\mathbf{F}_{i-1}) \in \mathbb{R}^{T_i \times D},
\end{equation}
where $T_i$ denotes the temporal length after downsampling. This process yields a sequence of temporally aligned feature maps ${\mathbf{F}_i}$, enabling the model to capture actions with varying temporal durations.

Finally, the Head layer is applied to predict temporal action instances from the temporal features $\{\mathbf{F}_i\}$. 
The head jointly performs action classification and temporal boundary regression, and the final prediction set is formulated as
\begin{equation}
\Psi_p = \mathrm{Head}(\{\mathbf{F}_i\})
      = \left\{ \phi_m = (t_s^m, t_e^m, c^m, s^m) \right\}_{m=1}^{M},
\end{equation}
where $m$ indexes the predicted action instances and $M$ denotes the total number of predictions. Predictions $\Psi_p$ from multiple temporal scales are aggregated and further refined by Non-Maximum Suppression (NMS)\cite{bodla2017soft} in post-processing, which removes redundant and highly overlapping proposals while preserving high-confidence detections. 

\begin{table*}[htbp]
\centering
\captionsetup{font=footnotesize}
\scriptsize
\caption{Comparison with previous methods on THUMOS14 and ActivityNet-1.3 Datasets.}
\label{tab:thumos-anet}
\begin{tabular}{l | c | c | c | cccccc | cccc}
\toprule
\multirow{2}{*}{\textbf{Method}} &
\multirow{2}{*}{\textbf{Backbone}} &
\multirow{2}{*}{\textbf{E2E}} &
\multirow{2}{*}{\textbf{Mem}} &
\multicolumn{6}{c|}{\textbf{THUMOS14}} &
\multicolumn{4}{c}{\textbf{ActivityNet-1.3}} \\
\cmidrule(lr){5-10}\cmidrule(lr){11-14}
& & & &
\textbf{0.3} & \textbf{0.4} & \textbf{0.5} & \textbf{0.6} & \textbf{0.7} & \textbf{Avg.} &
\textbf{0.5} & \textbf{0.75} & \textbf{0.95} & \textbf{Avg.} \\
\midrule
BSFTAL\cite{xu2025bfstal}             & InternVideo2\cite{chen2024video}           & \xmark & -- & 84.2 & 79.5 & 73.5 & 61.8 & 46.1 & 69.0 & 54.8  & 37.5  & 9.1   & 37.1  \\
CLTDR-GMG\cite{li2025temporal}        & InternVideo2\cite{chen2024video}  & \xmark & -- & 85.3 & 80.6 & 74.8 & 63.2 & 48.0 & 70.4 & 58.3  & 40.2  & 9.4   & 39.5  \\
VideoMamba\cite{li2025videomamba}    & InternVideo2\cite{chen2024video}  & \xmark & -- & 86.5 & 82.1 & 76.3 & 65.1 & 50.2 & 71.8 & 62.1  & 43.2  & 9.5   & 41.6  \\
ActionMamba~\cite{chen2024video}   & InternVideo2\cite{chen2024video}  & \xmark & -- & 87.1 & 82.7 & 76.5 & 65.6 & 49.9 & 72.3 & 62.2 &  37.1 & 9.6  & 41.7 \\
BDRC-Net\cite{fang2025boundary}        & InternVideo2\cite{chen2024video} & \xmark & -- & 87.3 & 83.7 & 78.1 & 67.4 & 53.1 & 73.9 & 59.6  & 38.9  & 9.8   & 41.9  \\
AdaTAD\cite{liu2023end}     & VideoMAEv2\cite{wang2023videomae} & \cmark & 29.9G & 86.8 & 82.3 & 76.1 & 65.3 & 49.9 & 72.1 & 59.8 & 42.0 & 9.7  & 40.8\\
AdaTAD++\cite{agrawal2025scaling} & VideoMAEv2\cite{wang2023videomae} & \cmark & 30.9G & 86.5 & 81.9 & 75.6 & 64.8 & 49.1 & 71.6 & 60.3 & 43.2 & 10.1 & 41.0 \\
MambaTAD\cite{lu2025mambatad}     & InternVideo2\cite{wang2022internvideo} & \cmark & 30.7G & 87.5 & 83.8 & 78.3 & 67.3 & 52.9 & 74.2 & 63.1 & 44.2 & 11.0  & 43.5 \\
\textbf{ESTF-SSM (Ours)} & VideoMAEv2\cite{wang2023videomae} & \cmark & 28.6G & \textbf{89.5} & \textbf{85.3} & \textbf{78.9} & \textbf{69.0} & \textbf{54.1} & \textbf{75.3} & \textbf{63.8} & \textbf{44.6} & \textbf{11.2} & \textbf{43.9} \\
\bottomrule
\end{tabular}
\end{table*}

\subsection{Efficient Spatial-Temporal Focal Adapter}

As illustrated in the lower-left part of Figure~\ref{fig:overview},  
the Efficient Spatial-Temporal Focal (ESTF) Adapter is inserted into selected pretrained backbone blocks as a lightweight adapter to enhance temporal modeling capability with minimal additional computation.

Given the intermediate feature representation $\mathbf{X}'_l \in \mathbb{R}^{N \times D}$ from the $l$-th pretrained block, the ESTF module first applies a channel down-projection to reduce computational cost:
\begin{equation}
\mathbf{Z} = \mathbf{X}'_l \mathbf{W}_{down}, \quad \mathbf{W}_{down} \in \mathbb{R}^{D \times D_r},
\end{equation}
where $D_r < D$ denotes the rank as the reduced channel dimension.
  
The projected features are then reshaped into spatio-temporal form $\mathbf{Z} \in \mathbb{R}^{T \times H \times W \times D_r}$.
The spatial branch focuses on local spatial context modeling by applying depthwise separable convolutions across spatial dimensions:
\begin{equation}
\mathbf{Z}'_{s} = \mathrm{DWConv}_{2d}(\mathbf{Z}),
\end{equation}
where $\mathbf{Z}'_{s}$ serves as the intermediate spatial feature. We combine it with the temporal features for fusion spatial-temporal modeling.
After upsampling, the temporal features $\mathbf{Z}'_{t}$ are added to obtain the result, which is further refined through a second 2D convolution to obtain the final spatial feature representation $\mathbf{Z}_{s}$:
\begin{equation}
\mathbf{Z}_{s} = \mathrm{DWConv}_{2d}(\mathbf{Z}'_{s} + \mathrm{UpSampling}(  \mathbf{Z}'_{t})).
\end{equation}
For the temporal branch, we first apply a 1D depthwise convolution. The resulting temporal features are then combined with the spatial features, followed by average pooling to reduce spatial dimensions. Finally, the SSM module is employed to capture global temporal dynamics, as formulated below:
\begin{equation}
\mathbf{Z}'_{t} = \big(\mathrm{DWConv}_{1d}(\mathbf{Z})\big), 
\end{equation}
\begin{equation}
\mathbf{Z}_{t} = \mathrm{TB\text{-}SSM}\big(\mathrm{AvgPool}(\mathbf{Z}'_{t} + \mathbf{Z}'_{s})\big).
\end{equation}
This design allows ESTF to efficiently encode both local and long-range temporal information without relying on quadratic self-attention.

The spatial and temporal features are then adaptively fused and projected back to the original channel dimension:
\begin{equation}
\mathcal{A}_l(\mathbf{X}'_l) = \big(\mathbf{Z}_{s} + \mathbf{Z}_{t} + \mathbf{Z}\big)\mathbf{W}_{up}.
\end{equation}
where $\mathbf{W}_{up} \in \mathbb{R}^{D_r \times D}$ denotes the channel up-projection. By decoupling spatial refinement and temporal focal modeling, the ESTF module achieves an effective balance between representation capacity and computational efficiency.

\subsection{Temporal Boundary-aware SSM}

The temporal modeling in ESTF is performed by our proposed Temporal Boundary-aware SSM (TB-SSM).
We build upon the basic Mamba~\cite{gu2023mamba}, which efficiently captures long-range dependencies through linear state updates. TB-SSM adopts independent state transition parameters for temporal forward and backward scans, while sharing the input-dependent projection parameters that generate $\mathbf{B}_t, \mathbf{C}_t$.

Given the input sequence $\mathbf{X}\in\mathbb{R}^{HW \times T \times D_r}$, we first apply layer normalization and an input projection:
\begin{equation}
\mathbf{X}^{f}, \mathbf{X}^{b} =\mathrm{Split}(\mathbf{W}_{in}(\mathrm{Norm}(\mathbf{X}))),
\end{equation}
where $\mathbf{W}_{in} \in \mathbb{R}^{D_r \times 2D_r}$ denotes the input projection of the TB-SSM, and two streams $\mathbf{X}^{f} \in \mathbb{R}^{HW \times T \times D_r}, \mathbf{X}^{b} \in \mathbb{R}^{HW \times T \times D_r}$ for forward and backward temporal scans split from $\mathbf{X} \in \mathbb{R}^{HW \times T \times 2D_r}$.
\begin{equation}
\mathbf{X}^{f}=\mathbf{X}, \quad
\mathbf{X}^{b}=\mathbf{X}.\mathrm{Flip}(),
\end{equation}
For the forward stream, TB-SSM performs selective state updates with an independent transition matrix $\mathbf{A}^{f}$:
\begin{align}
\mathbf{h}^{f}_t &= \mathbf{A}^{f} \mathbf{h}^{f}_{t-1} + \mathbf{B}_t \mathbf{X}^{f}_t, \\
\mathbf{Y}^{f}_t &= \mathbf{C}_t \mathbf{h}^{f}_t,
\end{align}
where $\mathbf{X}^{f}_t$ denotes the token at time step $t$.
For the backward stream, we apply the same update rule but with a different transition matrix $\mathbf{A}^{b}$:
\begin{align}
\mathbf{h}^{b}_t &= \mathbf{A}^{b} \mathbf{h}^{b}_{t-1} + \mathbf{B}_t \mathbf{X}^{b}_t, \\
\mathbf{Y}^{b}_t &= \mathbf{C}_t \mathbf{h}^{b}_t,
\end{align}
Finally, we concatenate the bidirectional outputs and apply an output projection to obtain the temporal representation:
\begin{equation}
\mathbf{Z}_{t} = \mathbf{W}_{out}(\mathbf{Y}^{f} \| \mathbf{Y}^{b}.\mathrm{Flip}()),
\end{equation}
where $\mathbf{W}_{out} \in \mathbb{R}^{2D_r \times D_r}$ denotes the output projection. 
This design enables direction-specific state dynamics for modeling asymmetric action boundaries, while keeping the overall temporal modeling linear in sequence length.

\begin{table}[htbp]
\centering
\scriptsize
\captionsetup{font=footnotesize}
\caption{Comparison with previous methods on Charades Dataset.}
\label{tab:charades}
\setlength{\tabcolsep}{4pt}
\renewcommand{\arraystretch}{1.1}
\begin{tabular}{l | c | c |ccc | c}
\toprule
\textbf{Method} & \textbf{Backbone} & \textbf{E2E} &
\textbf{0.2} & \textbf{0.5} & \textbf{0.7} & \textbf{Avg.} \\
\midrule
MS-TST\cite{dai2022ms}          & CLIP\cite{radford2021learning}   & \xmark   & 50.1 & 37.9 & 19.2& 31.9  \\
AAN\cite{dai2023attributes} & CLIP\cite{radford2021learning}   & \xmark   & 51.2 & 38.4 & 20.7 & 32.0 \\
MT-Temba~\cite{sinha2025ms}   & CLIP\cite{radford2021learning}   & \xmark   & 51.2 & 38.4 & 20.7 & 32.3 \\
AdaTAD~\cite{liu2023end}        & VideoMAEv2\cite{wang2023videomae}   & \cmark   & 53.7 & 42.9 & 26.7 & 37.5 \\
MambaTAD\cite{lu2025mambatad} & VideoMAEv2\cite{wang2023videomae} & \cmark & 54.1 & 43.2 & 26.6  & 37.7 \\
\textbf{ESTF-SSM (Ours)}    & VideoMAEv2\cite{wang2023videomae} & \cmark &
\textbf{57.2} & \textbf{45.5} & \textbf{27.1} & \textbf{38.9} \\
\bottomrule
\end{tabular}
\end{table}

\section{Experiments}
\subsection{Datasets and Metrics}
We conduct experiments on three datasets for temporal action detection. THUMOS14\cite{THUMOS14} contains 413 untrimmed videos with temporal annotations over 20 action categories, and is widely used for evaluating precise temporal localization. ActivityNet-1.3\cite{caba2015activitynet} is a larger benchmark with 19,994 videos and 200 categories, featuring diverse daily activities and long untrimmed videos. Charades\cite{sigurdsson2016hollywood} focuses on daily indoor activities with frequent action co-occurrence and complex temporal structures, providing a challenging testbed for long-range temporal reasoning.

We use the mean Average Precision (mAP) as the primary metric to evaluate the prediction results. The mAP is calculated by averaging the precision results across different temporal Intersection over Union (tIoU) thresholds. The tIoU is defined as the ratio of the intersection to the union of the predicted and ground-truth time intervals.

\subsection{Implementation Details}
We implemented our method on NVIDIA A100 GPUs, utilizing CUDA 12.8 within the PyTorch 2.8 environment. Throughout the model training process, we adopted the AdamW\cite{loshchilov2017decoupled} optimizer and employed a scheduler that integrates linear warm-up and cosine annealing to adjust the learning rate, which was set to 1e-4, with a batch size of 8. For video data, the sampling length of video frames was fixed at 768. Videos exceeding this length were subsampled to the maximum length, while those shorter were padded. Each frame was resized to a resolution of $224 \times 224$ pixels.

\subsection{Evaluation and Comparison}
We evaluate our method on the THUMOS14 and ActivityNet-1.3 benchmarks, comparing against both end-to-end (E2E) and non-E2E baselines. The quantitative results are presented in Table~\ref{tab:thumos-anet}. On THUMOS14, ESTF-SSM achieves superior average mAP, maintaining a consistent lead across all evaluated tIoU thresholds. This trend is particularly evident at high tIoU, indicating that our boundary-aware temporal modeling significantly refines boundary regression precision rather than merely improving coarse detection. On ActivityNet-1.3 dataset, ESTF-SSM demonstrates strong generalization ability by outperforming competitive baselines. The improvements are consistent across various overlap thresholds, verifying that our method effectively handles long-range temporal reasoning. Furthermore, regarding computational efficiency, our approach maintains a lower memory footprint compared to other E2E baselines, striking a favorable balance between performance and resource consumption.

We further extend our evaluation to the Charades dataset, as shown in Table~\ref{tab:charades}. ESTF-SSM consistently surpasses previous methods across all metrics. The performance gap is most distinct at tighter tIoU thresholds, which supports our motivation that modeling asymmetric boundary dynamics is crucial for preserving discriminative cues in scenarios involving complex activities and heavy background clutter.

\subsection{Qualitative Analyses}
\begin{figure}[htbp]
  \centering
  \captionsetup{font=scriptsize}
  \includegraphics[width=0.95\linewidth]{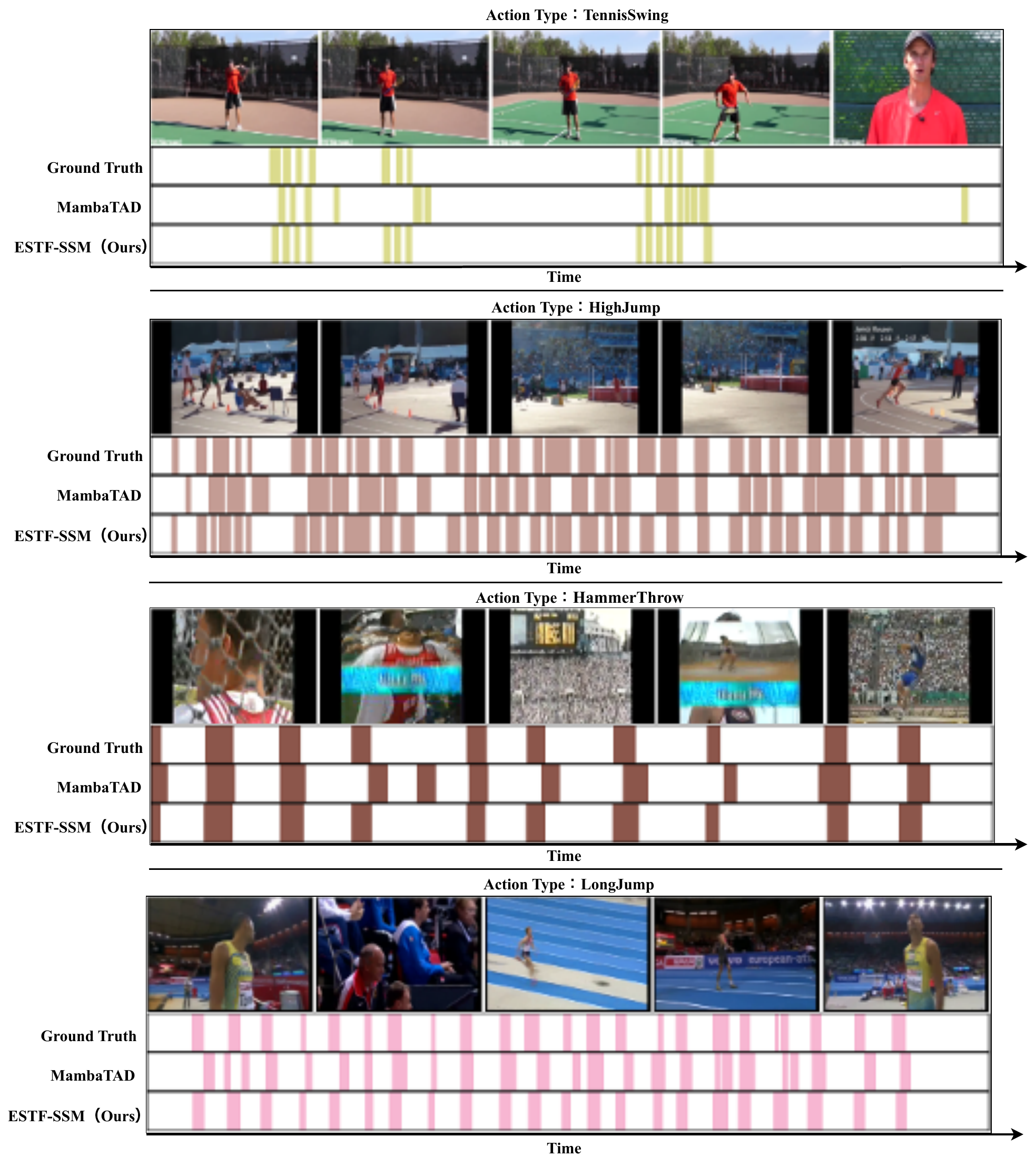}
  \caption{Qualitative results of our proposed method and previous method on THUMOS14 Dataset for Video Action Detection task.
  }
  \label{fig:visual_results}
\end{figure}

Figure~\ref{fig:visual_results} shows the qualitative visualization results for the video action detection task, where the horizontal axis represents time. The colored bars represent the predicted action segments with their corresponding categories, compared alongside the Ground Truth intervals representing the actual actions. Each action category is assigned a distinct color. Compared with the previous method\cite{lu2025mambatad}, our TB-SSM method demonstrates superior capability in recognizing action start and end boundaries.

\subsection{Ablation Studies}

We conduct comprehensive ablation studies on THUMOS14 and ActivityNet-1.3 to validate the contribution of each component in ESTF. The results are summarized in Table~\ref{tab:ablationA}. First, removing the temporal branch results in the most significant performance degradation, confirming that temporal modeling is the primary driver for long-range action reasoning. Second, discarding the spatial refinement branch also leads to consistent drops, suggesting that lightweight spatial processing remains essential for retaining localized appearance cues that aid in temporal boundary discrimination. Third, employing a simple parallel structure without the spatial-temporal fusion mechanism yields suboptimal results, indicating that explicit feature interaction is necessary to effectively couple spatial evidence with temporal dynamics. Finally, replacing TB-SSM with a variant lacking the proposed boundary-aware asymmetric state dynamics leads to inferior performance. This validates that direction-specific temporal state transitions are vital for capturing the intrinsic asymmetry between action onsets and offsets. Overall, the full model delivers the best results, demonstrating that these components contribute complementarily to accurate and robust temporal localization.

\begin{table}[t]
\centering
\scriptsize
\setlength{\tabcolsep}{3.2pt}
\renewcommand{\arraystretch}{1.12}
\captionsetup{font=footnotesize}
\caption{Ablation study on the effect of each component in the proposed method.}
\label{tab:ablationA}
\begin{tabular}{c c c c | c |c}
\toprule
\multicolumn{4}{c|}{\textbf{Components}} & \textbf{THUMOS14} & \textbf{ActivityNet-1.3} \\
\cmidrule(lr){1-4}\cmidrule(lr){5-5}\cmidrule(lr){6-6}
\textbf{Spatial} & \textbf{Temporal} & \textbf{S-T Fusion} & \textbf{TB-SSM} &
\textbf{mAP Avg.} & \textbf{mAP Avg.} \\
\midrule
\xmark & \cmark & \xmark & \cmark & 74.4 & 43.0 \\
\cmark & \xmark & \xmark & \xmark & 72.3 & 41.8 \\
\cmark & \cmark & \xmark & \cmark & 73.9 & 42.9 \\
\cmark & \cmark & \cmark & \xmark & 73.7 & 42.6 \\
\cmark & \cmark & \cmark & \cmark & \textbf{75.3} & \textbf{43.9} \\
\bottomrule
\end{tabular}
\end{table}

\begin{table}[htbp]
\centering
\scriptsize
\setlength{\tabcolsep}{3.8pt}
\renewcommand{\arraystretch}{1.12}
\captionsetup{font=footnotesize}
\caption{Ablation study on different temporal modeling strategies.}
\label{tab:ablation_temporal_strategy}
\begin{tabular}{l| c | c}
\toprule
\multirow{2}{*}{\textbf{Temporal Module Strategy}} &
\textbf{THUMOS14} & \textbf{ActivityNet-1.3} \\
\cmidrule(lr){2-3}
& \textbf{mAP Avg.} & \textbf{mAP Avg.} \\
\midrule
\textbf{w/o}         & 72.3 & 41.8 \\
Transformer\cite{vaswani2017attention} & 72.4 & 42.0 \\
VideoMamba\cite{li2025videomamba}     & 73.6 & 42.4 \\
ActionMamba\cite{chen2024video}       & 73.2 & 42.7 \\
TB-SSM(Ours)                        & \textbf{75.3} & \textbf{43.9} \\
\bottomrule
\end{tabular}
\end{table}

In Table~\ref{tab:ablation_temporal_strategy}, we further investigate various temporal modeling strategies. Our experiments show that Transformer-based modules offer only marginal gains compared to the baseline without temporal modeling, whereas SSM-based designs consistently yield better performance. Among them, linear-time state-space models like VideoMamba and ActionMamba improve detection accuracy, confirming their suitability for long untrimmed sequences. However, our TB-SSM achieves the highest performance, demonstrating that introducing boundary-aware asymmetric temporal dynamics provides a stronger inductive bias for temporal localization than conventional bidirectional or shared-parameter SSM designs.

\section{Conclusion}

We propose a novel temporal action detection framework, ESTF-SSM, which integrates the ESTF Adapter into a frozen video backbone. The ESTF module effectively combines the strengths of the SSM for long-range temporal modeling with efficient spatial refinement. By decoupling spatial and temporal processing and introducing the TB-SSM with asymmetric state dynamics, our approach successfully addresses the challenges of feature redundancy and boundary ambiguity in long video sequences. Extensive experiments on THUMOS14, ActivityNet-1.3, and Charades benchmarks demonstrate that our method significantly enhances localization performance and robustness compared to state-of-the-art methods. Future work will explore extending this efficient adaptation strategy to online action detection and multi-modal video understanding tasks.

{
\bibliographystyle{IEEEtran}
\bibliography{icme2026references}
}
\end{document}